\documentclass[sigconf,usenames,dvipsnames]{acmart}
\usepackage{fancyhdr}
\usepackage{enumitem}
\usepackage{amsmath}
\usepackage{amsfonts}
\usepackage[tight]{subfigure}
\usepackage{url}
\usepackage{mathtools}
\AtBeginDocument{%
  }
\usepackage{color}
\usepackage{float}
\usepackage{amsmath}
\usepackage{listings}
\usepackage[ruled,vlined]{algorithm2e}
\usepackage{amsmath}
\usepackage{amsfonts}
\usepackage{pifont}
\usepackage{hyperref}
\usepackage{breakurl}
\usepackage{float}
\usepackage[utf8]{inputenc}
\usepackage{subfigure}
\usepackage{graphicx}
\usepackage{graphicx}
\usepackage{textcomp}
\usepackage{xcolor}
\usepackage{url}

\definecolor{gitadd}{HTML}{00A64F}
\definecolor{gitdel}{HTML}{c94238}

\usepackage{upgreek}

\newcommand{\MYcomment}[1]{}

\newcommand{\MYnote}[1]{}

\newcounter{MYtablecntr}
\addtocounter{MYtablecntr}{1}

\newcommand{\MYlabel}{\small {$\bullet$}}

\newcounter{MYenumctrtwo}

\newcounter{MYenumctr}

\copyrightyear{2025}
\acmYear{2025}
\setcopyright{acmcopyright}
\acmConference[ICPP '25]{54th International Conference on Parallel Processing}{August 29-September 1, 2025}{San Diego, CA}
\acmDOI{10.1145/3750720.3758083}
\begin{document}

\title{Ortho-Fuse: Orthomosaic Generation for Sparse High-Resolution Crop Health Maps Through Intermediate Optical Flow Estimation}

\author{Rugved Katole} 
\affiliation{%
  \institution{The Ohio State University}
   \city{Columbus}
   \state{Ohio}  
  \country{United States}
}
\author{Christopher Stewart} 
\affiliation{%
  \institution{The Ohio State University}
   \city{Columbus}
   \state{Ohio}  
  \country{United States}
}

\begin{abstract}
\label{sect:abstract}
AI-driven crop health mapping systems offer substantial advantages over conventional monitoring approaches through accelerated data acquisition and cost reduction. However, widespread farmer adoption remains constrained by technical limitations in orthomosaic generation from sparse aerial imagery datasets. Traditional photogrammetric reconstruction requires 70-80\% inter-image overlap to establish sufficient feature correspondences for accurate geometric registration. AI-driven systems operating under resource-constrained conditions cannot consistently achieve these overlap thresholds, resulting in degraded reconstruction quality that undermines user confidence in autonomous monitoring technologies.
In this paper, we present Ortho-Fuse, an optical flow-based framework that enables the generation of a reliable orthomosaic with reduced overlap requirements. Our approach employs intermediate flow estimation to synthesize transitional imagery between consecutive aerial frames, artificially augmenting feature correspondences for improved geometric reconstruction. Experimental validation demonstrates a 20\% reduction in minimum overlap requirements. We further analyze adoption barriers in precision agriculture to identify pathways for enhanced integration of AI-driven monitoring systems. The code and dataset are available at  \href{https://rugvedkatole.github.io/OrthoFUSE/}{https://rugvedkatole.github.io/OrthoFUSE/}

\end{abstract}


\thispagestyle{plain}
\pagestyle{plain}

\maketitle
\section{Introduction}
\label{Sec: Introduction}
The integration of artificial intelligence with agricultural monitoring systems has emerged as a transformative approach to addressing the growing demands of digital agriculture and sustainable farming practices \cite{Ronald_Computer_2020}. AI-driven crop health mapping represents a critical advance in this domain, enabling farmers to monitor plant health, detect diseases early, and optimize resource allocation through automated aerial imagery analysis \cite{ouhami_computer_2021,maimaitijiang_crop_2020}. These systems typically rely on orthomosaic generation, i.e., the process of stitching together multiple aerial images to create seamless, geometrically corrected composite maps that provide comprehensive field coverage for subsequent analysis \cite{Ludwig_Quality_2020,ferrer_gonzalez_uav_2020}.

However, a fundamental technical challenge constrains the widespread adoption of AI-driven crop health mapping systems: the substantial data requirements for conventional orthomosaic generation. Traditional photogrammetric methods require 50-70\% overlap between adjacent images to ensure adequate feature matching and geometric consistency \cite{ferrer_gonzalez_uav_2020, Agisoft,barba_accuracy_2019}. This requirement stems from the need for sufficient common features between images to enable accurate image registration and seamless blending \cite{james_mitigating_2014,mills_global_2013,lin_blending_2016}. For agricultural applications, this translates to extensive flight missions covering 70-80\% of the field area to achieve the necessary overlap ratios. Consequently, this results in significant operational costs, extended flight times, and increased data processing complexity \cite{OpenDroneMaps, Pix4D, olson_review_2021}.

\begin{figure}[t]
    \centering
    \includegraphics[width=1\linewidth]{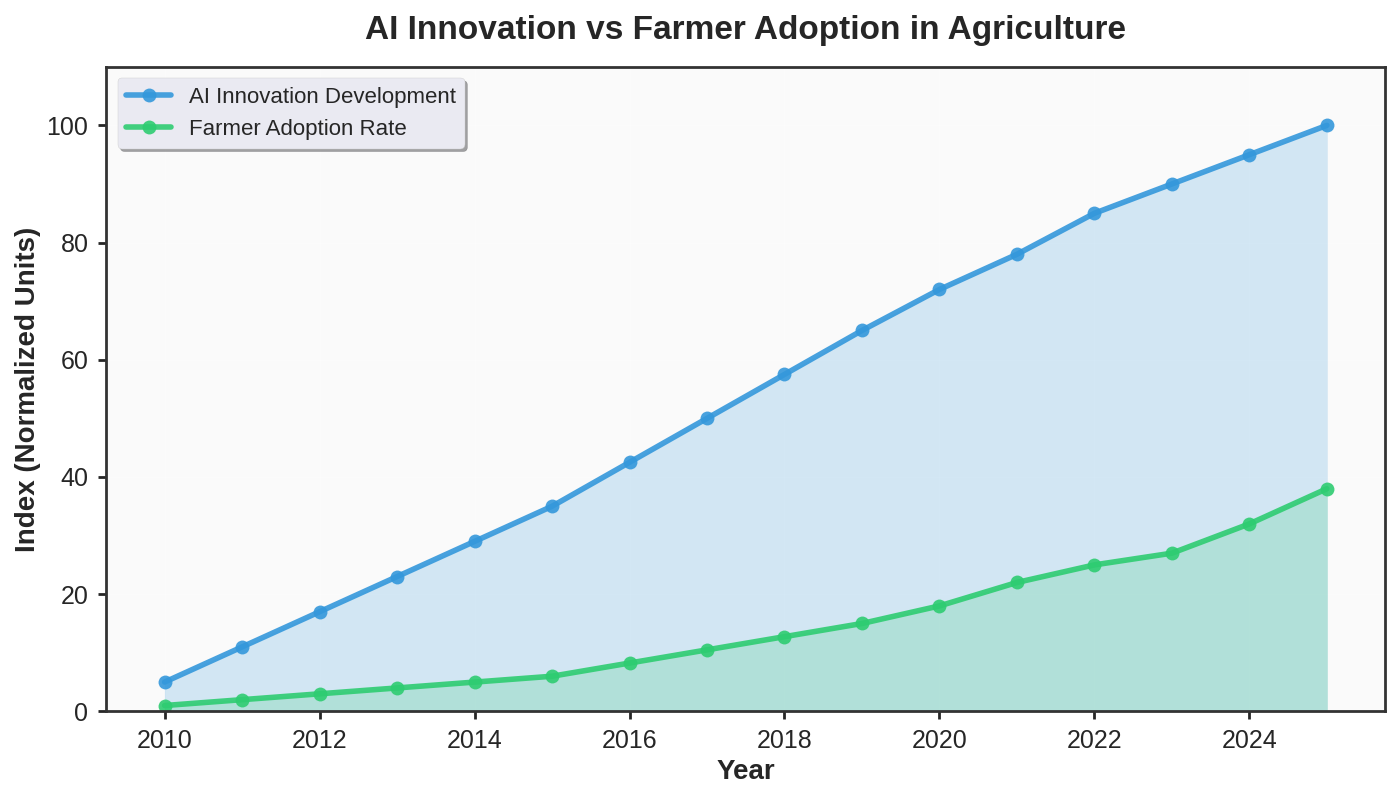}
    \caption{Trends in the number of AI innovations in Digital Agriculture and the number of new technologies adopted by farmers.$^{1}$}
    \label{fig:adoption_trend}
\end{figure}

The contradiction between AI systems' promise of efficiency and traditional orthomosaic requirements creates a critical bottleneck in practical deployment. While learning models for crop health assessment can operate with coverage as low as 20\% of the whole field and still achieve over 80\% accuracy in predicting the whole field health map \cite{katole2023multi,zhang_whole_field_2020}, the prerequisite for orthomosaic generation demands comprehensive coverage that negates these efficiency gains. This disconnect manifests particularly acutely in commercial agricultural operations where cost-effectiveness and operational efficiency directly impact adoption rates \cite{McFadden2023, Balasundram_ROLE_2023}.

Current software solutions for orthomosaic generation, including industry-standard platforms such as Pix4D \cite{Pix4Dmapper2025}, Agisoft Metashape \cite{AgisoftMetashape2025}, OpenDroneMaps \cite{OpenDroneMap2025}, and DJI Terra \cite{DJITerra2025}, exhibit significant performance degradation when confronted with sparse data scenarios \cite{brach2019accuracy}. The diminished feature detection rates in sparse datasets result in poor image alignment, visible seams, geometric distortions, and reduced overall map quality that renders subsequent AI analysis unreliable \cite{Shahi_Recent_2023}. This limitation is particularly pronounced in agricultural environments where repetitive crop patterns and seasonal variations further complicate feature detection and matching processes \cite{chiu2020agriculture}.

To address this fundamental limitation, this paper introduces Ortho-Fuse, an intermediate optical flow estimation-based paradigm for creating high-quality orthomosaics with reduced overlap requirements. The proposed system leverages a real-time intermediate flow estimation (RIFE) model \cite{huang2022real} for temporal interpolation of synthetic images between two aerial images, thereby enhancing overlapping features between sparse data samples. By generating contextually appropriate synthetic imagery that preserves spatial relationships and agricultural features, Ortho-Fuse enables the production of high-quality orthomosaics from reduced data collection requirements while maintaining accuracy within acceptable thresholds for downstream crop health analysis. The significance of this approach extends beyond technical innovation to address practical deployment challenges that have historically limited the adoption of AI-driven agricultural monitoring systems. By reducing data collection requirements by up to 20\% while maintaining analytical accuracy, Ortho-Fuse offers a pathway to economically viable precision agriculture systems.

The innovation-adoption disparity illustrated in Figure \ref{fig:adoption_trend}\footnote{Data Sources: \cite{GAO2024,MarketsandMarkets2023,GrandViewResearch2023,AFE2022}. The graph represents a projection for reference and does not depict actual ground truth values.} reveals fundamental methodological limitations that constrain the practical deployment of digital agriculture technologies. Current AI-driven health mapping systems demonstrate remarkable predictive accuracy exceeding 80\% through selective field scouting of merely 20\% coverage. However, translating these sparse health predictions into comprehensive orthomosaics presents formidable technical challenges. The inherent spatial discontinuities and non-overlapping feature distributions within limited-coverage health maps create substantial algorithmic complexities for orthomosaic reconstruction. This fundamental disconnect between high-accuracy sparse health prediction and comprehensive spatial visualization constitutes a critical barrier to practical deployment, directly contributing to the persistent innovation-adoption gap observed in precision agriculture technologies.

In this paper, our contributions are as follows:
\begin{itemize}
    \item \textbf{Ortho-Fuse:} An optical flow-based paradigm for generating orthomosaics from sparse aerial image data with reduced overlap requirements.
    
    \item \textbf{Challenges and Future Directions:} We discuss the barriers to adoption of AI methods and identify research directions for bridging the gap between innovation and practical deployment.
\end{itemize}

The remainder of the paper is organized as follows. Section \ref{sec:Related Works} discusses related works and highlights the problem. Section \ref{sec:methods} explains the core methodology and its implementation. Section \ref{sec:results} showcases the results obtained through our framework. Finally, Section \ref{sec:Conclusion} discusses the conclusion and future directions.

\section{Related Works}
\label{sec:Related Works}

\subsection{Traditional Photogrammetry and UAV Processing Pipelines}
\label{subsec:traditional_photogrammetry}
The field of orthomosaic generation has evolved significantly since its early foundations in photogrammetry, with traditional approaches emphasizing geometric accuracy and seamless image blending through high-overlap requirements. \cite{Ludwig_Quality_2020} demonstrated that conventional UAV processing pipelines require 75\% front overlap and 60-75\% side overlap to achieve acceptable geometric accuracy, with forest applications demanding even higher overlap ratios exceeding 90\%. These requirements stem from the fundamental dependence on feature detection and matching algorithms, where sufficient common features between adjacent images are essential for robust image registration and bundle adjustment processes \cite{FENG2023107650}. The ISPRS Journal of Photogrammetry and Remote Sensing has published extensive research on seamline detection and optimization strategies for orthomosaic generation, highlighting the critical role of blending zones and hierarchical structures in addressing color discontinuities and geometric distortions \cite{mills_global_2013, lin_blending_2016}.

\subsection{Contemporary Software Solutions and Performance Limitations}
\label{subsec:contemporary_software}
Contemporary software solutions have attempted to address some limitations of traditional approaches but remain constrained by fundamental overlap requirements. Agisoft Metashape and Pix4D, representing industry-standard platforms, demonstrate consistent performance degradation when processing sparse datasets due to insufficient feature correspondences \cite{ferrer_gonzalez_uav_2020,barba_accuracy_2019}. Recent work by \cite{brach2019accuracy} compared six photogrammetric software packages across forest conditions, revealing average RMSE values of 1.24m for raw orthophotos with precision improvements to approximately 0.2m only after extensive post-processing. This performance degradation becomes particularly pronounced in agricultural environments where crop patterns create additional challenges for feature detection algorithms \cite{Bauer2019, Bouguettaya2022}.

\subsection{AI-Driven Agricultural Monitoring Systems}
\label{subsec:ai_agricultural_monitoring}
The emergence of AI-driven approaches in agricultural monitoring has created new possibilities for crop health assessment while simultaneously highlighting the limitations of traditional orthomosaic generation methods. \cite{ouhami_computer_2021} demonstrated that machine learning models for crop disease detection can achieve 95-99\% accuracy using CNN architectures, with ResNet34 achieving 99.67\% accuracy on standard datasets. However, these systems remain dependent on high-quality orthomosaic inputs, creating a bottleneck where the promise of AI efficiency is negated by extensive data collection requirements \cite{agronomy14091975}. 
The disconnect between AI capabilities and traditional orthomosaic requirements has been identified as a critical factor limiting the practical deployment of precision agriculture systems \cite{Sishodia2020,waltz2025cyberinfrastructure}.

\subsection{Computer Vision Applications in Agricultural Contexts}
\label{subsec:computer_vision_agriculture}
Computer vision applications in agriculture have advanced significantly, with comprehensive surveys revealing the dominance of convolutional neural networks in crop monitoring applications. \cite{Balasubramanian_2020} provided extensive coverage of deep learning applications in plant phenotyping, emphasizing the potential for automated agricultural analysis. The Agriculture-Vision dataset, containing 94,986 high-quality aerial images from 3,432 farmlands, addresses challenges in extreme annotation sparsity while demonstrating the feasibility of large-scale agricultural image analysis \cite{chiu2020agriculture}. However, these advances in computer vision capabilities have not been matched by corresponding improvements in orthomosaic generation efficiency, creating a fundamental mismatch between data collection requirements and analytical capabilities \cite{Wu2022}.

\subsection{Sparse Data Reconstruction Techniques}
\label{subsec:sparse_data_reconstruction}
Sparse data reconstruction techniques have emerged as a promising approach to address data limitation challenges across various computer vision domains. Recent developments in neural radiance fields (NeRF) have demonstrated effective sparse-view reconstruction capabilities, with \cite{du20241stplacesolutioniccv} achieving first-place performance in ICCV 2023 competitions using Pixel-NeRF with depth supervision. CNN-based approaches utilizing Voronoi tessellation have shown promise for reconstructing global spatial fields from sparse sensor data, suggesting potential applications in agricultural monitoring scenarios \cite{sunderhaft2024deeplearningimprovementssparse}. However, these techniques have not been systematically adapted to address the specific challenges of orthomosaic generation in agricultural contexts, where domain-specific constraints and requirements differ significantly from general computer vision applications \cite{na2024uforecongeneralizablesparseviewsurface}.

\subsection{Generative AI Approaches for Image Synthesis}
\label{subsec:generative_ai_approaches}
Generative AI approaches have shown remarkable potential for addressing sparse data challenges through synthetic image generation and data augmentation techniques. Diffusion models, in particular, have demonstrated superior performance in generating high-quality synthetic imagery while maintaining spatial consistency and feature preservation \cite{chae2025aptadaptivepersonalizedtraining,ifriqi2025improvedconditioningmechanismspretraining}. MetaEarth, a generative foundation model for global-scale remote sensing image generation, represents a significant advancement in agricultural and remote sensing applications \cite{yu2024metaearthgenerativefoundationmodel}. However, existing generative approaches have not been specifically designed to address the unique requirements of orthomosaic generation, where synthetic images must maintain precise spatial relationships and feature correspondences necessary for successful image registration \cite{tang2024crsdiffcontrollableremotesensing}.

\subsection{Digital Agriculture Adoption Challenges}
\label{subsec:adoption_challenges}
The challenges of digital agriculture adoption have been extensively documented, with multiple studies identifying cost, trust, complexity, and technical limitations as primary barriers to widespread implementation. \cite{Balasundram_ROLE_2023} identified significant barriers, including limited access to technology, high implementation costs, and insufficient data quality, as critical factors limiting adoption. 
\cite{romero2023ai} noted that practitioners may distrust AI-driven recommendations, insisting on manual confirmation before adoption.
Research by \cite{chaterji2020artificialintelligencedigitalagriculture} emphasized the need for a comprehensive understanding of on-farm decision-making processes and privacy-preserving frameworks for data collaboration. Zhang et al.~\cite{zhang2022assessing} highlighted the need for AI tailored to specific decision-making needs (e.g., spraying versus manual verification). 
However, tailoring AI to specific contexts has led to a plethora of models. Practitioners now struggle with the complexity of digital agriculture solutions where distinct, bespoke solutions are required for different crops~\cite{yang2020adaptive,sarkar2023assessment}, data modalities~\cite{ockerman2023reflection}, and reinforcement learning contexts~\cite{boubin2022marble}.
For orthomosaic generation, cost is a primary concern, but sampling imposes trust concerns. Our goal is to provide a simple and trustworthy method that can complement low-cost scouting methods~\cite{zhang_whole_field_2020,katole2023multi}.

\subsection{Feature Detection Challenges in Agricultural Aerial Imagery}
\label{subsec:feature_detection_challenges}
Feature detection in aerial imagery presents unique challenges in agricultural contexts due to repetitive crop patterns, seasonal variations, and changing environmental conditions. Deep learning approaches have shown promise in addressing these challenges, with CNN architectures demonstrating robust performance in crop disease detection and monitoring applications \cite{Picon2019, Kamilaris2018}. However, the fundamental limitation remains that traditional orthomosaic generation software relies on feature detection algorithms that degrade significantly with sparse data, creating a bottleneck that limits the practical application of advanced AI techniques in agricultural monitoring \cite{Luo2024}. The development of specialized approaches that can maintain robust feature detection and image registration performance under sparse data conditions represents a critical research need for enabling widespread AI-driven agricultural monitoring systems \cite{Ghazal2024}.

\subsection{Research Gap and Synthesis}
\label{subsec:research_gap}
The convergence of these research areas—traditional photogrammetry, AI-driven agricultural monitoring, sparse data reconstruction, and generative AI—suggests significant potential for novel approaches that can address the fundamental limitations of current orthomosaic generation methods. While existing literature has advanced individual components of this challenge, the integration of generative AI techniques specifically designed for agricultural orthomosaic generation from sparse data remains largely unexplored.

\section{Methodology}
\label{sec:methods}

\begin{figure}[t]
    \centering
    \includegraphics[width=3in]{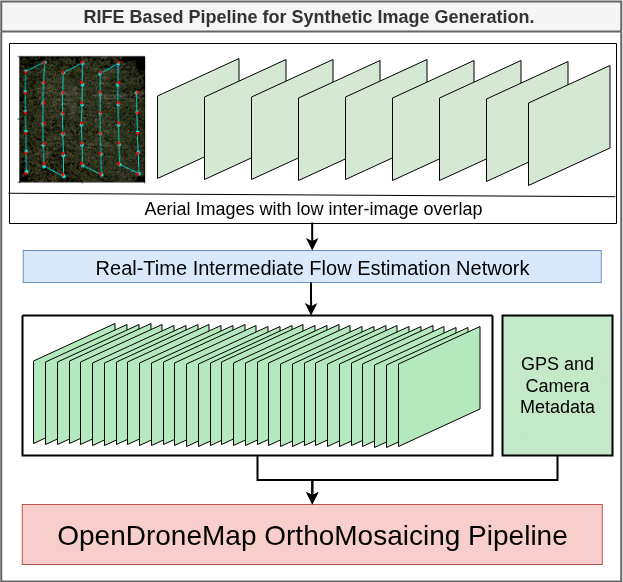}
    \caption{An Intermediate Flow estimation-based Pipeline Orthomosaic generation with lower overlapping features.}
    \label{fig:Rife}
\end{figure}

The conventional approach to crop health assessment requires exhaustive aerial scouting to capture comprehensive field coverage. The collected data is subsequently processed using software platforms like OpenDroneMaps and Pix4D to generate orthomosaics for visualization. However, U.S. farmlands spanning hundreds of acres make exhaustive scouting laborious and cumbersome, necessitating multiple flights to collect complete field data. Traditional orthomosaic generation requires 70-75\% image overlap for optimal results, which extends flight duration as only 20-25\% new information is captured with each image. \textbf{Ortho-Fuse} addresses this inefficiency by generating intermediate images between consecutive frames to enhance orthomosaic quality with reduced overlap requirements.

\begin{figure*}[t]
    \centering
    \includegraphics[width=\linewidth]{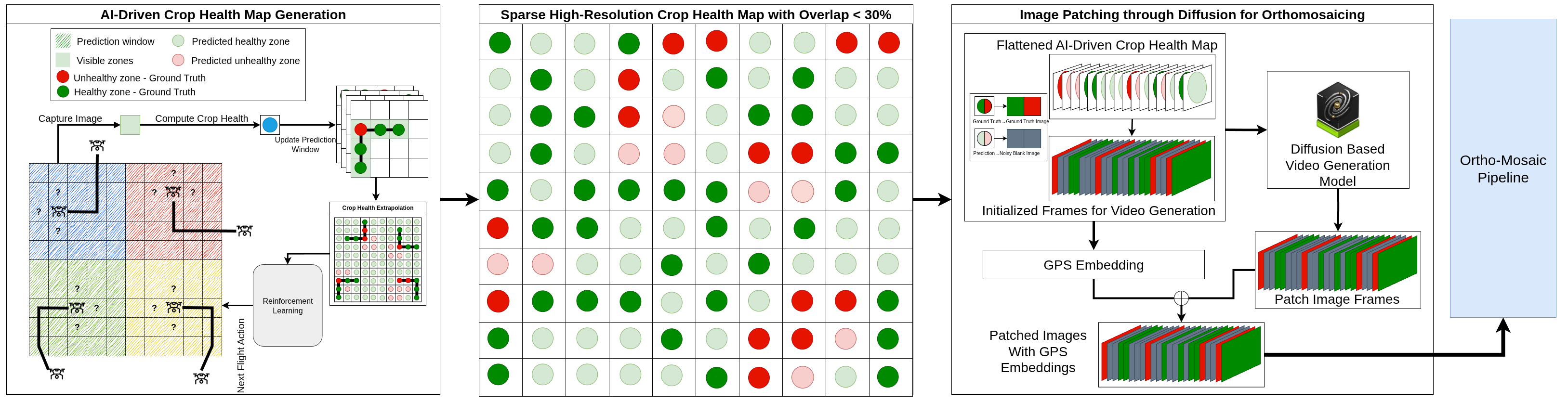}
    \caption{Proposed Framework for Diffusion-based motion estimation and synthetic frame generation to enable Orthomosaicing with AI-Driven Sparse crop health maps. AI-Driven Health maps source: \cite{katole2023multi}}
    \label{fig: overview}
\end{figure*}

Figure \ref{fig:Rife} illustrates the Ortho-FUSE framework. The process begins with UAV image dataset collection, where inter-image overlap can vary between 25-50\%, with results varying accordingly. The collected data is processed through the Real-Time Flow Estimation (RIFE) network \cite{huang2022real} to generate intermediate frames through temporal interpolation between existing frames.

RIFE employs a neural architecture called IFNet that directly estimates intermediate optical flows (Ft→0, Ft→1) and fusion masks from consecutive video frames. This approach circumvents traditional multi-stage flow reversal methodologies through end-to-end learning paradigms. The framework incorporates a teacher-student training strategy where a teacher model with ground truth intermediate frame access guides student network optimization, enabling accurate flow estimation without dependency on pre-trained optical flow models. The system synthesizes intermediate frames through backward warping operations on input frames using predicted flows, achieving real-time video interpolation with superior computational efficiency compared to conventional flow-based methodologies. RIFE leverages deterministic motion-guided synthesis with explicit temporal constraints from consecutive video frames. Unlike generative adversarial networks and diffusion models that rely on stochastic sampling processes and iterative denoising procedures lacking inherent temporal coherence guarantees, RIFE ensures consistent motion preservation and computational determinism. Critically, RIFE operates as a pre-trained model requiring no domain-specific retraining for agricultural applications, enabling direct deployment on aerial crop imagery without additional training overhead. This temporal-aware architectural paradigm establishes RIFE as an optimal choice for intermediate frame generation applications requiring real-time performance, motion consistency, and resource-efficient deployment.

The generated intermediate frames lack essential metadata including GPS coordinates and camera parameters required for orthomosaic generation. These parameters determine the reconstruction technique and final results. We address this by linearly interpolating GPS coordinates between frames while maintaining the same camera parameters as the original images. With newly generated frames and embedded metadata, this enhanced dataset is processed through the orthomosaicing pipeline using OpenDroneMaps (ODM). ODM subsequently generates high-quality orthomosaics from sparse datasets. Section \ref{sec:results} compares accuracy between original datasets and synthetically generated images.

\subsection{Limitations and Future Directions}

Contemporary flow-estimation methodologies demonstrate robust performance for sequential frame synthesis, ensuring temporally coherent transitions between consecutive image representations. However, these frameworks exhibit degraded accuracy as inter-frame semantic similarity diminishes, fundamentally constraining their capacity to synthesize complex object motion trajectories between disparate visual representations.

The proposed intermediate flow estimation approach demonstrates optimal performance within agricultural monitoring applications due to substantial visual homogeneity and consistent environmental patterns across temporal datasets. Conversely, when deployed on heterogeneous datasets exhibiting asymmetric visual characteristics and irregular motion patterns, the framework experiences performance degradation due to its reliance on optical flow estimation rather than comprehensive object motion modeling.

This architectural limitation restricts practical deployment to specialized application domains characterized by repetitive visual patterns and minimal inter-frame variation, precluding broader applicability in dynamic environments requiring robust object tracking capabilities. Furthermore, the methodology's dependence on motion continuity assumptions limits effectiveness in scenarios involving occlusion, sudden illumination changes, or discontinuous object trajectories. Future research directions should explore hybrid approaches incorporating semantic understanding and object-level motion representation.

\subsection{Adoption Challenges and Technical Barriers}

Beyond algorithmic constraints, orthomosaic generation complexities constitute a primary barrier limiting widespread adoption of AI-driven agricultural systems. Current adoption remains at only 27\% of U.S. farms, despite demonstrated yield improvements of 15-30\% \cite{GAO2024}.

Technical barriers in orthomosaic processing manifest through exponential computational scaling, requiring 65-145 minutes for 1,030-image datasets and multiple days for large-scale operations (77,000+ images), with memory consumption reaching 50+ GB RAM. Agricultural environments present unique challenges, including repetitive crop patterns that confuse feature matching algorithms, resulting in 30-50\% initial outlier ratios and 5-15\% image incorporation failure rates \cite{ZHU2023120525}. Structure-from-Motion (SfM) algorithms struggle with homogeneous crop surfaces, requiring 75\%+ image overlap and producing geometric accuracy limitations of 2-5 cm horizontal and 3-4 cm vertical without ground control points \cite{FENG2023107650}.

Systemic adoption challenges compound these technical barriers. Economic constraints dominate farmer decision-making, with 52\% citing high upfront costs and 40\% uncertain about return on investment \cite{Fiocco_mcinsey}. Infrastructure limitations further restrict deployment—only 26\% of rural farms possess adequate broadband connectivity essential for real-time AI processing \cite{trimble_precision_2022}. The aging farmer demographic (average 58 years) combined with 38\% reporting insufficient technical expertise creates additional adoption friction \cite{blasch_farmer_2020,finszter_challenges_2024,noauthor_ai_2024}.

Given farmers' limited technical expertise, they rely on intuitive methods like orthomosaics that provide visual cues for understanding overall field health. Future research directions should prioritize diffusion-based orthomosaic generation methodologies as illustrated in Figure \ref{fig: overview}, where image patching through diffusion models enables robust orthomosaic synthesis from sparse high-resolution crop health maps with reduced overlap requirements (<30\%). This approach potentially addresses fundamental SfM limitations through GPS-embedded patch reconstruction, offering computational efficiency improvements while maintaining geometric accuracy.

\subsection{Advanced Generative Approaches}

Advanced diffusion-based video generation models, exemplified by Nvidia Cosmos \cite{nvidia2025cosmosworldfoundationmodel}, demonstrate sufficient computational capacity to synthesize coherent video sequences from singular textual or visual inputs. These diffusion architectures can be systematically fine-tuned to model drone trajectory dynamics while capturing temporal-spatial ground movement transitions between discrete image acquisitions.

As illustrated in Figure \ref{fig: overview}, this methodology enables robust orthomosaic reconstruction from highly sparse and irregularly distributed datasets collected through AI-driven scouting algorithms. Interpolated video sequences generated from sparsely sampled imagery serve as comprehensive training datasets for orthomosaic synthesis pipelines. The fundamental merit of this approach lies in its dual capacity to reduce technological barriers for agricultural practitioners while simultaneously enhancing computational resource utilization efficiency in precision agriculture applications, thereby addressing both accessibility constraints and processing optimization challenges inherent in contemporary digital farming methodologies.

The convergence of GPU acceleration, edge computing, and novel generative approaches suggests imminent transformation in agricultural AI adoption, contingent upon resolving orthomosaic generation bottlenecks through automated processing pipelines and real-time quality assessment systems.

\section{Results and Evaluation}
\label{sec:results}

This section presents a comprehensive evaluation of the Ortho-Fuse framework through comparative analysis of orthomosaic generation using three distinct approaches: (1) traditional reconstruction from original images with standard overlap, (2) reconstruction using exclusively RIFE-generated synthetic intermediate frames, and (3) hybrid reconstruction combining both original and synthetic images. Our evaluation demonstrates that synthetic and hybrid approaches achieve superior reconstruction quality while preserving crop health and analytical accuracy.

\begin{figure}
    \centering
    \includegraphics[width=0.75\linewidth]{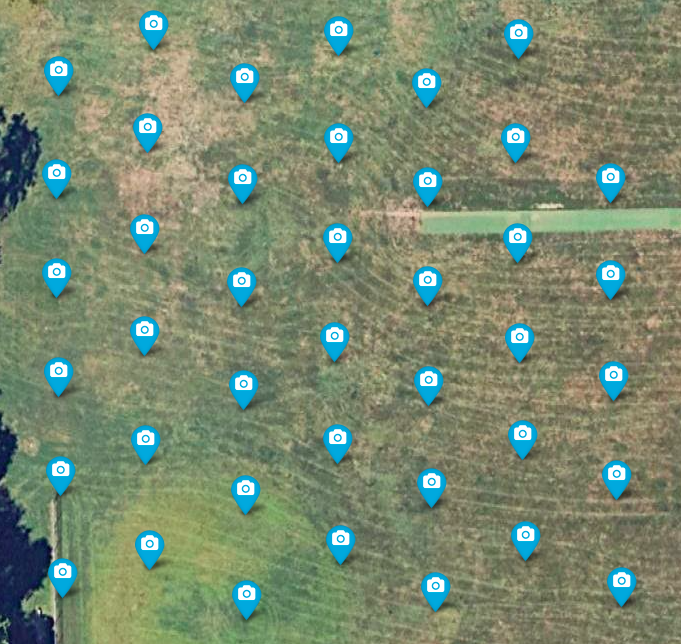}
    \caption{Ground Control Points (GCP) distribution and flight path for data collection}
    \label{fig:GCP}
\end{figure}

\subsection{Dataset and Experimental Setup}
\begin{figure*}[t]
    \centering
    \includegraphics[scale=0.30]{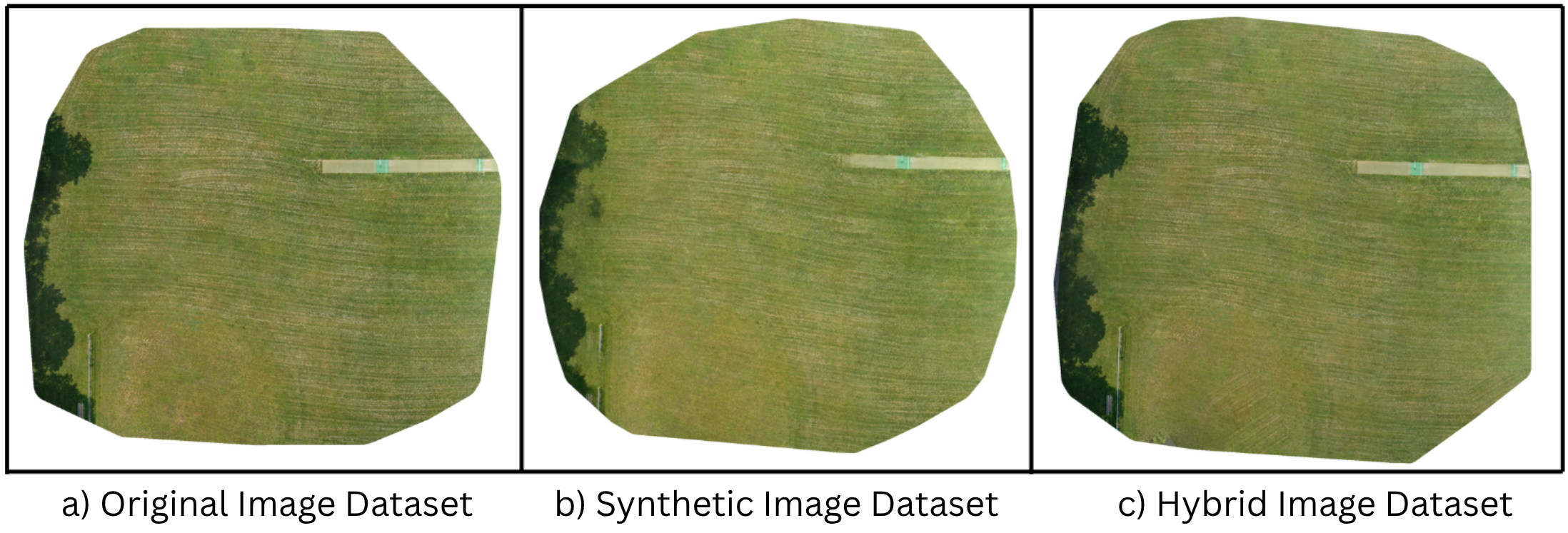}
    \caption{Comparative orthomosaic quality: (a) Original 50\% overlap, (b) Synthetic frames only, (c) Hybrid approach}
    \label{fig:orthomosaic_comparison}
\end{figure*}

The evaluation was conducted using two aerial imagery datasets collected from agricultural fields with controlled 50\% side and front overlap, as illustrated in Figure \ref{fig:GCP}. Data was collected using the Parrot Anafi drone at 15 meters height from ground level, following the flight path shown in \ref{fig:GCP}. For every pair of images in the original dataset, we generated three synthetic images, creating a pseudo-overlap of 87.5\%. 

The experimental design follows a three-tier comparative framework:

\begin{itemize}
    \item \textbf{Baseline (Original):} Traditional orthomosaic generation using only original aerial images with 50\% overlap
    \item \textbf{Synthetic:} Orthomosaic generation using exclusively RIFE-generated intermediate frames
    \item \textbf{Hybrid:} Combined approach utilizing both original images and synthetic intermediate frames
\end{itemize}

All orthomosaics were processed using OpenDroneMap Web ODM version with identical parameter configurations to ensure comparative validity. Ground control points (GCPs) were established across each test field to enable quantitative accuracy assessment, as depicted in Figure \ref{fig:GCP}.

\subsection{Orthomosaic Visual Quality Evaluation}

Figure \ref{fig:orthomosaic_comparison} illustrates representative sections of orthomosaics generated using each approach. The synthetic and hybrid methods demonstrate improved seamline integration and reduced artifacts compared to the baseline reconstruction. Additionally, the average Ground Sample Distance (GSD) for the original dataset, synthetic, and hybrid data was measured as 1.55 cm, 1.49 cm, and 1.47 cm, respectively. These measurements indicate that synthetic and hybrid datasets provide enhanced granularity in images, resulting in superior orthophoto quality.

\subsection{Crop Health Analysis}

\begin{figure}[t]
    \centering
    \includegraphics[width=1\linewidth]{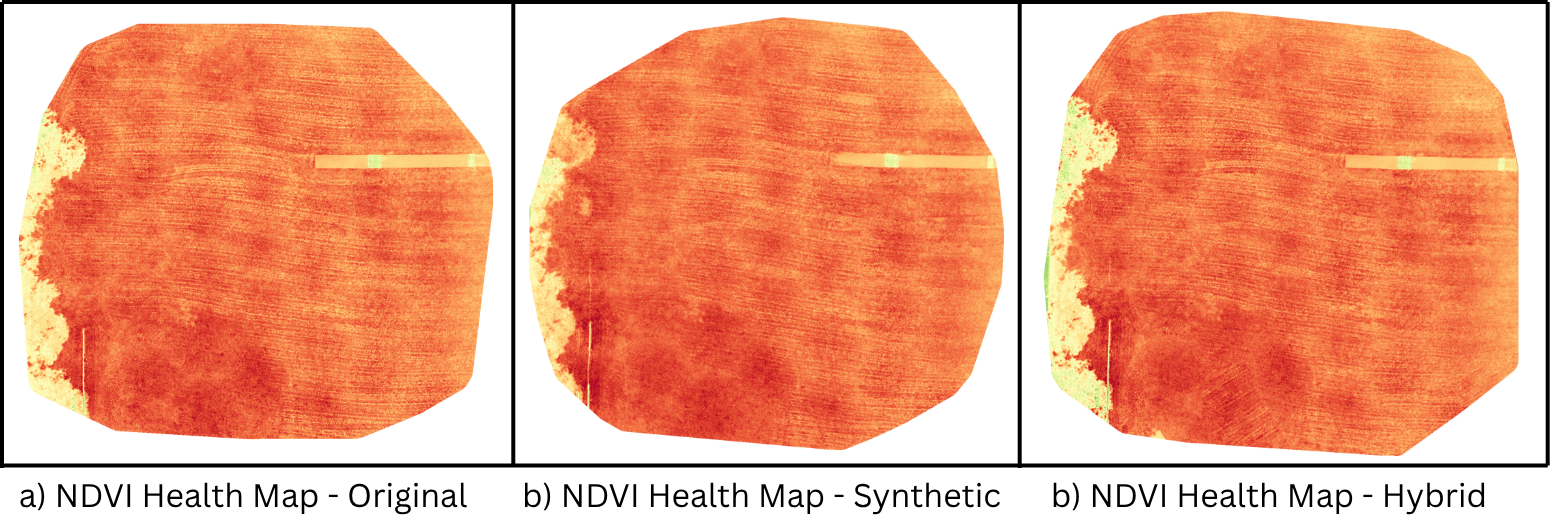}
    \caption{NDVI crop health maps: (a) Original orthomosaic NDVI, (b) Synthetic orthomosaic NDVI, (c) Hybrid orthomosaic NDVI}
    \label{fig:ndvi_health_maps}
\end{figure}

To validate that synthetic frame integration preserves agricultural analytical accuracy, we conducted crop health assessments using NDVI analysis across all three orthomosaic variants. Figure \ref{fig:ndvi_health_maps} presents NDVI-derived crop health visualizations computed from all three orthomosaic variants. The color-coded health maps enable direct comparison of vegetation indices across reconstruction approaches, demonstrating consistent agricultural analytical capabilities.

\section{Conclusion and Future Work}
\label{sec:Conclusion}
This paper presents Ortho-Fuse, an optical flow-based framework that addresses the critical bottleneck of generating high-quality orthomosaics from sparse aerial imagery in AI-driven precision agriculture. Our approach successfully demonstrates reliable orthomosaic reconstruction with significantly reduced overlap requirements through RIFE-based synthetic frame generation.

The experimental validation reveals three key achievements. First, Ortho-Fuse enables orthomosaic generation with only 50\% inter-image overlap while maintaining reconstruction quality comparable to traditional methods requiring 70-80\% overlap, representing a 20\% reduction in minimum overlap requirements. Second, synthetic and hybrid reconstruction approaches achieve superior visual quality metrics, with Ground Sample Distance improvements from 1.55 cm (original) to 1.47 cm (hybrid). Third, NDVI-based crop health assessment validates that synthetic frame integration preserves agricultural analytical accuracy essential for farmer adoption.

The framework directly addresses the innovation-adoption disparity in digital agriculture by enabling cost-effective orthomosaic generation that supports farmer decision-making processes. Integration with existing photogrammetric pipelines ensures compatibility with industry-standard software platforms, facilitating practical implementation without comprehensive system overhauls.

Despite its advantages, the optical flow-based approach is limited primarily to agricultural environments with visual homogeneity and consistent patterns. Performance degradation occurs with heterogeneous datasets exhibiting asymmetric visual characteristics or irregular motion patterns. Ortho-Fuse represents a significant step toward resolving the disconnect between AI capabilities and practical deployment requirements in Digital agriculture. We also highlight the challenges and future directions for bridging this gap. By reducing the disparities between innovation and deployment, we can make better agricultural monitoring systems that are economically viable and practically deployable at scale, supporting sustainable agricultural practices and enhanced food security.

\noindent {\bf Acknowledgments:} 
This work was funded by ACCESS Computing Startup Grant CIS220074, NSF Grants OAC-2112606, and the Ohio Soybean Council.




{
\renewcommand{\bibfont}{\normalsize}
\bibliographystyle{ACM-reference-format} 
\bibliography{ref}
}

\end{document}